\documentclass{article}

\usepackage[final]{ml4molecules_2022}

\usepackage[utf8]{inputenc} 
\usepackage[T1]{fontenc}    
\usepackage{hyperref}       
\usepackage{url}            
\usepackage{booktabs}       
\usepackage{amsfonts}       
\usepackage{nicefrac}       
\usepackage{microtype}      
\usepackage{xcolor}         
\usepackage{graphicx}
\usepackage{subcaption}

\title{Dynamic Molecular Graph-based Implementation for Biophysical Properties Prediction}

\author{%
  Carter Knutson \\
  Pacific Northwest National Laboratory \\
  Washington, WA 99354 \\
  \texttt{carter.knutson@pnnl.gov} \\
   \And
   Gihan Panapitiya \\
   Pacific Northwest National Laboratory \\
   Washington, WA 99354 \\
   \texttt{gihan.panapitiya@pnnl.gov} \\
   \And
   Rohith Varikoti \\
   Pacific Northwest National Laboratory \\
   Washington, WA 99354 \\
   \texttt{Rohith.Varikoti@pnnl.gov} \\
   \AND
   Neeraj Kumar \\
   Pacific Northwest National Laboratory \\
   Washington, WA 99354 \\
   \texttt{Neeraj.Kumar@pnnl.gov} \\
}

\begin{document}

\maketitle

\begin{abstract}
   Neural Networks (GNNs) have revolutionized the molecular discovery to understand patterns and identify unknown features that can aid in predicting biophysical properties and protein-ligand interactions. However, current models typically rely on 2-dimensional molecular representations as input, and while utilization of 2\textbackslash3-dimensional structural data has gained deserved traction in recent years as many of these models are still limited to static graph representations. We propose a novel approach based on the transformer model utilizing GNNs for characterizing dynamic features of protein-ligand interactions. Our message passing transformer pre-trains on a set of molecular dynamic data based off of physics-based simulations to learn coordinate construction and make binding probability and affinity predictions as a downstream task. Through extensive testing we compare our results with the existing models, our MDA-PLI model was able to outperform the molecular interaction prediction models with an RMSE of 1.2958. The geometric encodings enabled by our transformer architecture and the addition of time series data add a new dimensionality to this form of research.
\end{abstract}

\section{Introduction}
Machine learning (ML) has greatly advanced protein structure prediction and molecular design processes, however, functional and reliable molecular representation remains a fundamental problem. In this context, molecular structural constraint is a major contributing factor to this challenge. Understanding protein-ligand interactions (PLIs) is vital for drug candidate design and discovery. Many machine learning models rely on 2-dimensional sequence information such as simplified molecular-input line-entry system strings (SMILES) or protein sequences, or 3-dimensional information haphazardly extrapolated from the fore mentioned 2D strings \citep{torng2019graph,ozturk2018deepdta, ragoza2017protein}.  Reliance on 2D data creates an environment in which designed or discovered molecules are syntactically aligned with their predecessors, but completely lack semantic validity. 3D data extrapolated from a purely 2D representation is greatly lacking in spatial accuracy, a core component to characterize PLIs. 3D data is expensive and time consuming to obtain through traditional or computational methods. Therefore it is critical to utilize available data to the fullest extent.  

Graph Neural Networks (GNNs) introduced graph data capabilities for the purpose of deep learning \citep{Velickovic_2017, gao2021topology, kipf2016semi}. Graph based molecular design and discovery models have gained significant traction in recent years. These representations are a natural choice for molecular representation in machine learning as atoms easily translate to nodes and edges and have shown improvement when compared to earlier methods and data formats \citep{duvenaud2015convolutional, shlomi2020graph}. Top performing PLI focused GNNs, including those that manage 3D molecular data are still limited to translational invariant, and static models \citep{lim2019predicting, knutson2022decoding,gonczarek2016learning}. This greatly hinders molecular graph representation and learning by depriving the model of important physical\textbackslash geometric properties.

Geometric and dynamic functionalities are a fairly new and natural development of the GNN. There are multiple ways to represent the geometric data associated with each atom's 3D coordinate set. In this context, \citet{ashby2021geometric} created a dynamic model in which edge prediction is coupled with the Euclidean distance between atoms. Atoms represented by their Cartesian coordinates enable \citet{xu2022geodiff}'s model to predict molecular conformations. More recently, message passing transformer models have gained popularity for the inclusion of geometric data within molecular learning models \citet{liu2021spherical, schutt2017schnet, vignac2020building, brandstetter2021geometric}.

In this work, we propose a novel approach to PLI exploration by utilizing message passing transformers to learn on a limited set of dynamic time series PLI data derived from MD simulations so called the MD-Assisted-PLI(MDA-PLI). This model is unique in its ability in order to make biophysical property predictions on general static data downstream after pre-training on a limited MD dataset. Our approach was inspired by  recent pre-train to finetune pipelines \citep{devlin2018bert, liu2021pre, wu2022pre}. We combine the geometric benefits afforded to message passing transformers with the consideration of time-series data to create a truly dynamic molecular representation model.

\section{Methods}
\subsection{Architecture}

\begin{figure}[h]
\centering
    \includegraphics[scale=0.3]{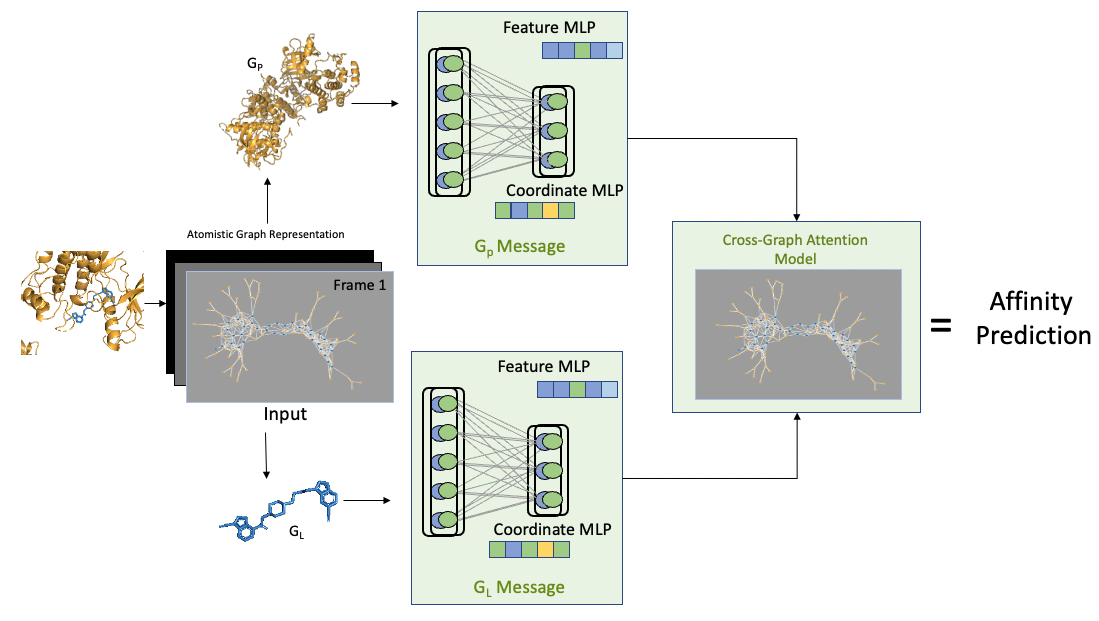}
\caption{Illustration of the pre-training phase and general downstream logic. Molecular dynamic snapshots are converted into time series graphs. The protein and ligand graph messages of each complex frame are fed into a GNN layer with gated cross graph attention that completes the pre-training stage. Binding affinity prediction is made downstream based on the same logic flow. }
\label{arch}
\end{figure}

The ligand and protein structures are represented in terms of two separate graphs, $G_{L}(V_{L},E_{L})$ and $G_{P}(V_{P},E_{P})$, where $V$ and $E$ correspond to vertices and edges of the graph. Each node is associated with 116 features. 
Our  model is based on the equivariant graph neural network architectures \cite{Satorras_2021} and \cite{Ganea_2021}. The model consists of initial node embedding message updates for ligand and protein graphs within a cross-graph attention mechanism that conducts the coordinate updates as shown in Figure \ref{arch}. 

The node features are updated using the message passing mechanism. During the individual graph encoding phase, the message that is sent from a source node to a target node is constructed as shown in Equation \ref{eq:message1}. The node features of the target ($h_{i}^{l}$) and source ($h_{j}^{l}$), and the squared relative distance between them ($||x_{i} - x_{j}||^{2}$) are first concatennated and then transformed using a multilayer perceptron $\phi_{e}$.

\begin{equation}
   m_{i,j} = \phi_{e}( h_{i}^{l}, h_{j}^{l} , ||x_{i} - x_{j}||^{2} ), ~\forall(i,j) \in G_{P}, G_{L}
   \label{eq:message1}
\end{equation}

What is unique in our model is the use of multiple aggregation methods. We use sum, mean and max aggregation to form three messages,


\begin{equation}
    h_{i}^{sum} = \sum_{j \in  N(i)} m_{i,j},  ~~ h_{i}^{mean} = \frac{1}{|N(i)|}\sum_{j \in  N(i)} m_{i,j}, ~~ h_{i}^{max} =  max(m_{i,j}).
    \label{eq:message_aggr}
\end{equation}

Then, we form the final message $M_{i}$ by linearly transforming the concatenation of $h_{i}^{sum}, h_{i}^{mean}$ and $h_{i}^{max}$. That is $h_{i}$ is given by,

\begin{equation}
    M_{i} = \phi_{aggr}(h_{i}^{sum}, h_{i}^{mean}, h_{i}^{max})
    \label{eq:message_final}
\end{equation}

The ligand and protein messages are used in the graph interaction layer to update the respective node embeddings. For each structure, we update the coordinates according to Equation \ref{eq:coors2} after transforming $m_{i,j}$ using the multilayer perceptron $\phi_{x}$.

\begin{equation}
x_{i}^{l+1} = x_{i}^{l} + \sum_{j \in N(i) } ||x_{i}-x_{j}||^{2}\phi_{x}(m_{i,j})
   \label{eq:coors2}
\end{equation}

The remaining steps of the mathematical formulation of our architecture are given in the Appendix.



 \subsection{Data}
Pre-training is conducted on graphs created from the MD simulation frames of 33 unique but diverse carefully selected protein-ligand complexes from the RCSB \citet{10.1093/nar/gkaa1038}. Simulations were run for 50ns with snapshots captured every 10ps resulting in roughly 5,000 frames per target. Targets were divided into a standard 80-10-10 training, validation, and test set split at random. Some samples were unable to be run or processed in their entirety, please see the Table\ref{tab:data_table} in the Appendix for a more explicit breakdown of samples included in each pre-training set. Downstream fine-tuning and affinity prediction tasks were performed on 7,695 experimental samples pulled from PDBBind2016 in the same division splits as the pre-training sets. For computational efficiency protein graphs are cropped only to incorporate the pocket in which the ligand is bound through a k-hop proximity method. Pre-training and downstream run on the same hyperparemeters for 30 epochs with early stopping at a patience of 25.

\section{Results and Discussion}

As a first step, we evaluate our model on the widely used PDBBind2016 dataset, specially prepared for our model to assess the general performance. Distribution of the affinity values in the train and test sets is shown in Figure \ref{fig:hist}, and the predicted versus actual affinity values in Figure \ref{fig:pt_predictions} in the Appendix. The model generally finds it challenging to make predictions for structures associated with low affinity values as indicated in the residual values shown in Figure \ref{fig:residuals} also located in the Appendix. 

In Table \ref{tbl:res}, we present our results in comparison to the experiments on similar datasets reported by \citet{wu2022pre} and \citet{feng2021attention}. The top portion shows the results from two sequence based models: DeepDTA by \citet{ozturk2018deepdta} and DeepAffinity by \citet{karimi2019deepaffinity}.  Structure based models are listed afterwards with our results in the final row. As shown in Table \ref{tbl:res}, we are able to achieve top-of-the-line results when compared to other structure based models. 

\begin{table*}[th]
\centering
\begin{tabular}{llll}
\toprule
Model & RMSE &    Pearson &   Spearman \\
\midrule
DeepDTA\citep{ozturk2018deepdta} & 1.565 & 0.573 & 0.574 \\
DeepAffinity\citep{karimi2019deepaffinity} & 1.893 & 0.415 & 0.436 \\
\midrule
3DCNN\citep{townshend2020atom3d} & 1.429$\pm$ 0.042 & 0.541$\pm$ 0.029 & 0.532$\pm$ 0.033 \\
IEConv\citep{hermosilla2020intrinsic} & 1.554$\pm$ 0.016 & 0.414$\pm$ 0.053 & 0.428$\pm$ 0.032 \\
ProtMD\citep{wu2022pre} LP\footnotemark[1] & 1.413$\pm$ 0.032 & 0.572$\pm$ 0.047 & 0.569$\pm$ 0.051\\
ProtMD\citep{wu2022pre} FT\footnotemark[2] & 1.367$\pm$ 0.014 & 0.601$\pm$ 0.036 & 0.587$\pm$ 0.042\\
GCAT\citep{feng2021attention} & 1.382 & 0.585 & 0.592 \\
Pafnucy\citep{stepniewska2018development} & 1.489 & 0.539 & 0.537\\
\midrule
\textbf{MDA-PLI without pre-train (Current Work)} & 1.3462 & 0.5955  & 0.5444\\
\textbf{MDA-PLI with pre-train (Current Work)} & 1.2958 & 0.6181  & 0.5698\\

\bottomrule
\end{tabular}
\caption{PDBBind Comparison Results of RMSE, Pearson correlation, and Spearman correlation}
\label{tbl:res}
\end{table*}

Multiple observations can be made from the presented results. DeepDTA and DeepAffinity exhibit the worst performance as the only sequence based models present in the table. Performance increase produced by structure base models is clear. While many of the reported structure based models have additional years of technological advancement afforded to them, \citet{stepniewska2018development} et al. is a prime comparison in the argument for the inclusion of 3D structural data.

Many of the models reported are still limited to static protein-ligand representations, even when based on 3D crystal structure. In this context, ProtMD by \citet{wu2022pre} and GCAT by \citet{feng2021attention} can be noted as the top performing publications that inspired our current work. Like ProtMD we also utilize MD data in a pre-training stage before the downstream affinity prediction. Both previous models preform predictions through some form of geometrically informed GNN, in the case of GCAT with an attention mechanism. We also implement cross-graph attention consideration to bolster our affinity predictions and lay the foundation to further investigate the mechanisms behind protein-ligand binding affinity.

In Figure \ref{fig:binned}, we show how the Mean Absolute Error (MAE) for different affinity regions vary. The blue colored numbers in \ref{fig:binned} are the number of training data points for each affinity region. We can see that the lack of training data points is a major factor that affects the prediction accuracy. However, pretraining has been beneficial to improve the predictions for affinity regions for which we have relatively a small number of training data points. The model trained from scratch performs slightly better than the the pretrained model in the high affinity region.



\begin{figure*}[!th]
     \centering
    \begin{subfigure}[t]{0.48\textwidth}
         \centering
         \includegraphics[scale=0.45]{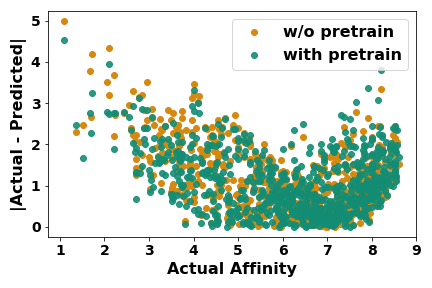}
         \caption{}
         \label{fig:residuals}
     \end{subfigure}
         \begin{subfigure}[t]{0.48\textwidth}
         \centering
         \includegraphics[scale=0.45]{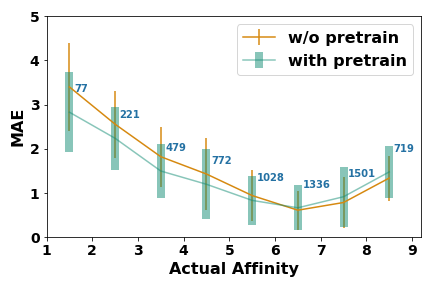}
         \caption{}
         \label{fig:binned}
     \end{subfigure}

    \caption{(a) Residuals of the predictions. (b) Binned Mean Absolute Error by affinity. Number of training data points for each affinity region is shown in blue.} 
        \label{fig:residuals-comp}
\end{figure*}




\section{Conclusion}

In this work, we devised a model that used MD data to pre-train a model based on the Equivariant Graph Neural Networks in order to improve the structure-property relationship of the protein-ligand interactions. We employ multiple aggregation along with cross graph attention to achieve high affinity prediction accuracies compared to previous models. In particular, our pretraining has been instrumental in improving the predictions for affinity regions with low amounts of training data.

\section{Acknowledgements}
This work was supported by the Laboratory Directed Research Funding (LDRD), Mathematics of Artificial Reasoning for Science (MARS) Initiative, at the Pacific Northwest National Laboratory (PNNL). PNNL is a multiprogram national laboratory operated by Battelle for the DOE under Contract DEAC05-76RLO 1830. This research used computational resources provided by Research Computing at the Pacific Northwest National Laboratory.

\bibliography{neurips_2022}{}
\bibliographystyle{neurips_2022}

\appendix

\section{Appendix}

\subsection{Mathematics}

We employ cross graph attention to account for the interaction between the neighboring atoms of the ligand and the protein. Only the nearest inter-graph neighbors such that the distance between them is less than $th_{dist}$ is considered to find attention coefficient. The attention mechanism we use here is what is introduced by \citet{Velickovic_2017}. The cross-graph message between nodes $i$ in the ligand and nodes $j$ in the protein (or vice versa) is given by, 

\begin{equation}
    \mu_{i,j} = a_{i,j}\textbf{W}h_{j}^{enc}, ~~ ||x'_{i} - x'_{j} || < th_{dist}
\end{equation}
The attention coefficients $a_{i,j}$ are defined as,

 \begin{equation}
  a_{i,j} = \frac{exp(LeakyRelU(\textbf{a}[ \textbf{W}h_{i}, \textbf{W}h_{j} ] ))}{\sum_{k \in N(i)} exp(LeakyRelU(\textbf{a}[\textbf{W}h_{i}, \textbf{W}h_{k} ]))},
 \end{equation}
 
where \textbf{a} and \textbf{W} are weight matrices. The cross-graph message are aggregated by adding according to, 

\begin{equation}
\mu_{i} = \sum_{j \in G_{P}} \mu_{i,j}~~ \forall i \in G_{L} ~~ and ~~ \mu_{i} = \sum_{j \in G_{L}} \mu_{i,j}~~ \forall i \in G_{P}.
\end{equation}

The final node feature update can now be carried out using linear transformation function $\Phi^{n}$ according to Equation \ref{eq:final-update}.

\begin{equation}
    h_{i}^{l+1} = h_{i}^{enc} + \Phi^{n}(h_{i}^{enc}, M_{i}, \mu_{i} ).
    \label{eq:final-update}
\end{equation}

During pre-training, the model learns to predict the coordinates of the next time step, based on those of the current time step. Mean squared loss is used as the loss function. The finetuning task is to predict protein-ligand binding affinity. We sum the final node embeddings to arrive at a representation for the whole structure. These representations for the ligand and the protein are concatenated and transformed to the dimensions of the finetuning target using a multi-layer perceptron.

\subsection{Data}
An explicit breakdown of pre-training datasets and sample counts.
\begin{table}[h]

\begin{center}
\begin{tabular}{lll}
\toprule
Set  & Targets & Total Samples \\ 
\midrule
Train         &27          &116,116 \\
Test             &3         &15,000 \\
Validation             &3         &15,000 \\
\bottomrule
\end{tabular}
\end{center}
\caption{Pre-training data.}
\label{tab:data_table}
\end{table} 

\break 

\subsection{Results}
\begin{figure*}[!h]
     \centering
    \begin{subfigure}[b]{0.48\textwidth}
         \centering
         \includegraphics[width=\textwidth]{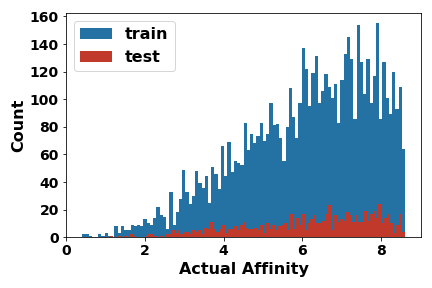}
         \caption{}
         \label{fig:hist}
    \end{subfigure}
    \begin{subfigure}[b]{0.48\textwidth}
         \centering
         \includegraphics[width=\textwidth]{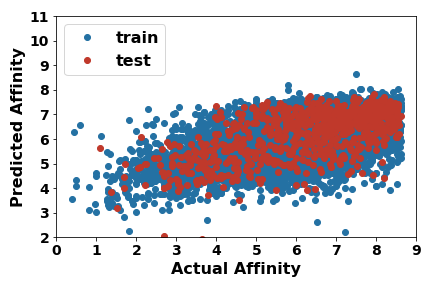}
         \caption{}
         \label{fig:pt_predictions}
     \end{subfigure}

    \caption{(a) Distribution of affinity values in train and test sets. (b) Train and test set predictions using the pre-trained model.}
        \label{fig:res}
\end{figure*}



\end{document}